\documentclass[11pt,letterpaper]{article}
\usepackage{ijcnlp2017}
\usepackage{times}
\usepackage{latexsym}
\usepackage[noend]{algpseudocode}
\usepackage{algorithm}
\usepackage{graphicx,caption}
\usepackage{float}
\usepackage{url,amsmath,setspace,amssymb,caption}

\ijcnlpfinalcopy

\title{Finding Dominant User Utterances And System Responses in Conversations}

\author{Dhiraj Madan  \and Sachindra Joshi \\
  IBM Research Labs\\
  New Delhi,  India\\
  \tt  {\{dhimadan,jsachind@in.ibm.com\} } \\}
\date\today{}

\begin{document}

\maketitle

\begin{abstract}
There are several dialog frameworks which allow manual specification of intents and rule based dialog flow. The rule based framework provides good control to dialog designers at the expense of being more time consuming and laborious. The job of a dialog designer can be reduced if we could identify pairs of user intents and corresponding responses automatically from prior conversations between users and agents. In this paper we propose an approach to find these frequent user utterances (which serve as examples for intents) and corresponding agent responses. We propose a novel SimCluster algorithm that extends standard K-means algorithm to simultaneously cluster user utterances and agent utterances by taking their adjacency  information into account. The method also aligns these clusters to provide pairs of intents and response groups. We compare our results with those produced by using simple K-means clustering on a real dataset and observe upto 10\% absolute improvement in F1-scores. 
Through our experiments on synthetic dataset, we show that our algorithm gains more advantage over K-means algorithm when the data has large variance.

\end{abstract}
\section{Introduction}

There are several existing works that focus on modelling conversation using prior human to human conversational data \cite{Gas13,You13,Hen14}. \cite{Hi11} models the conversation from pairs of consecutive tweets. Deep learning based approaches have also been used to model the dialog in an end to end manner \cite{Vin15,Ser15}. Memory networks have been used by Bordes et al \shortcite{Bor16} to model goal based dialog conversations. More recently, deep reinforcement learning models have been used for generating interactive and coherent dialogs \cite{Li16} and negotiation dialogs \cite{Lew17}.

\begin{figure*}[h!]
\begin{center}

  \includegraphics[width=0.9\linewidth]{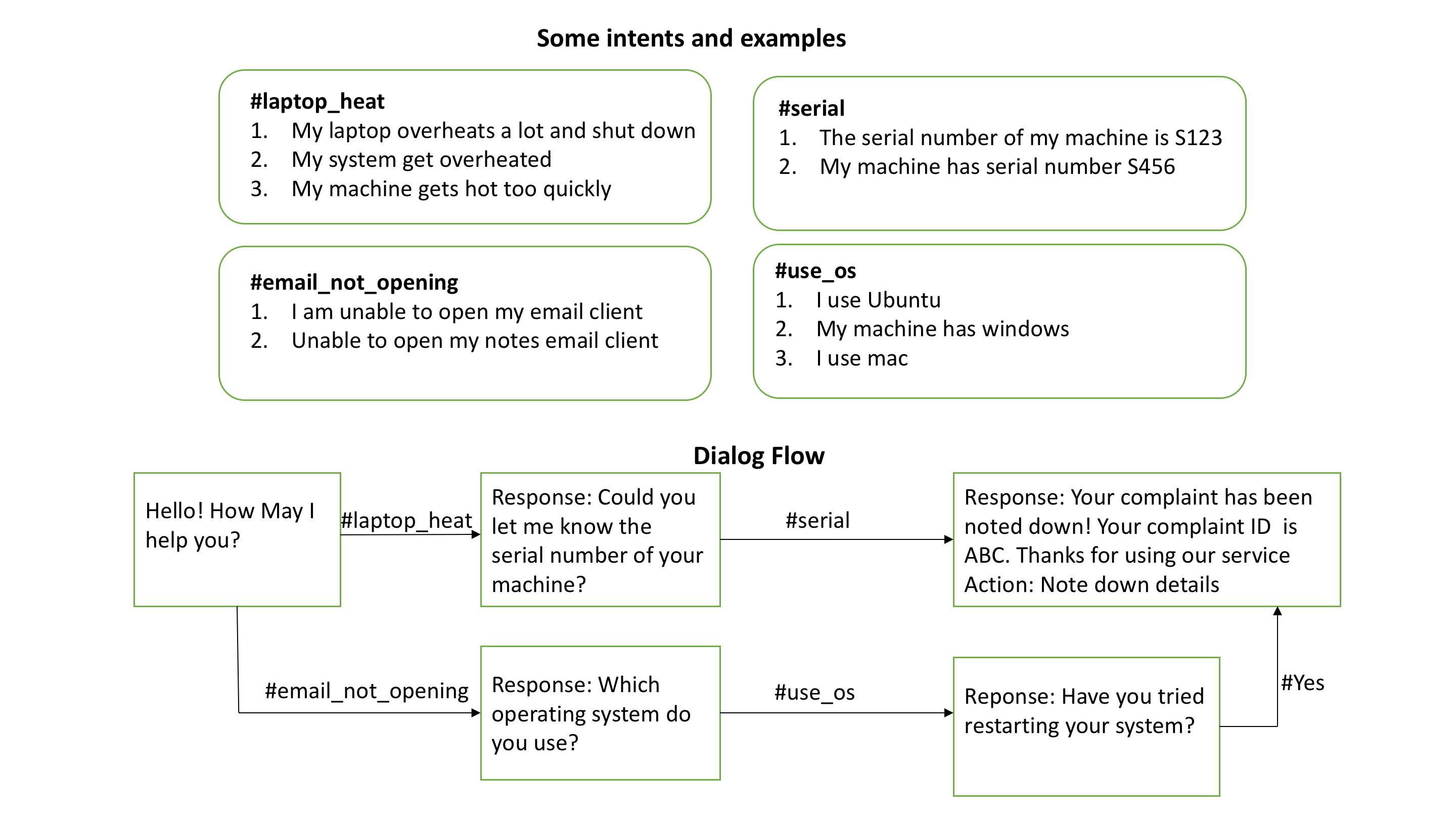}
  \caption{Some intents and dialog flow}
  \label{fig:intents_entities_flow}
\end{center}

\end{figure*}

Industry on the other hand has focused on building frameworks that allow manual specification of dialog models such as api.ai\footnote{https://api.ai/}, Watson Conversational Services\footnote{https://www.ibm.com/watson/developercloud/conversation.html}, and Microsoft Bot framework\footnote{https://dev.botframework.com}. 
These frameworks provide ways to specify {\it intents}, and a {\it dialog flow}. 
The user utterances are mapped to intents that are passed to a dialog flow manager. The dialog manager generates a response and updates the dialog state.  See Figure \ref{fig:intents_entities_flow} for an example of some intents  and a dialog flow in a technical support domain. The dialog flow shows that when a user expresses an intent of {\it \# laptop\textunderscore heat}, then the system should respond with an utterance {\it ``Could you let me know the serial number of your machine ''}. The designer needs to specify intents (for example {\it \# laptop\textunderscore heat}, {\it \# email\textunderscore not\textunderscore opening}) and also provide corresponding system responses in the dialog flow.
This way of specifying a dialog model using intents and corresponding system responses manually is more popular in industry than a data driven approach as it makes dialog model easy to interpret and debug as well as provides a better control to a dialog designer. 
However, this is very time consuming and laborious and thus involves huge costs. 

\begin{figure*}
\begin{center}
  \includegraphics[width=\linewidth]{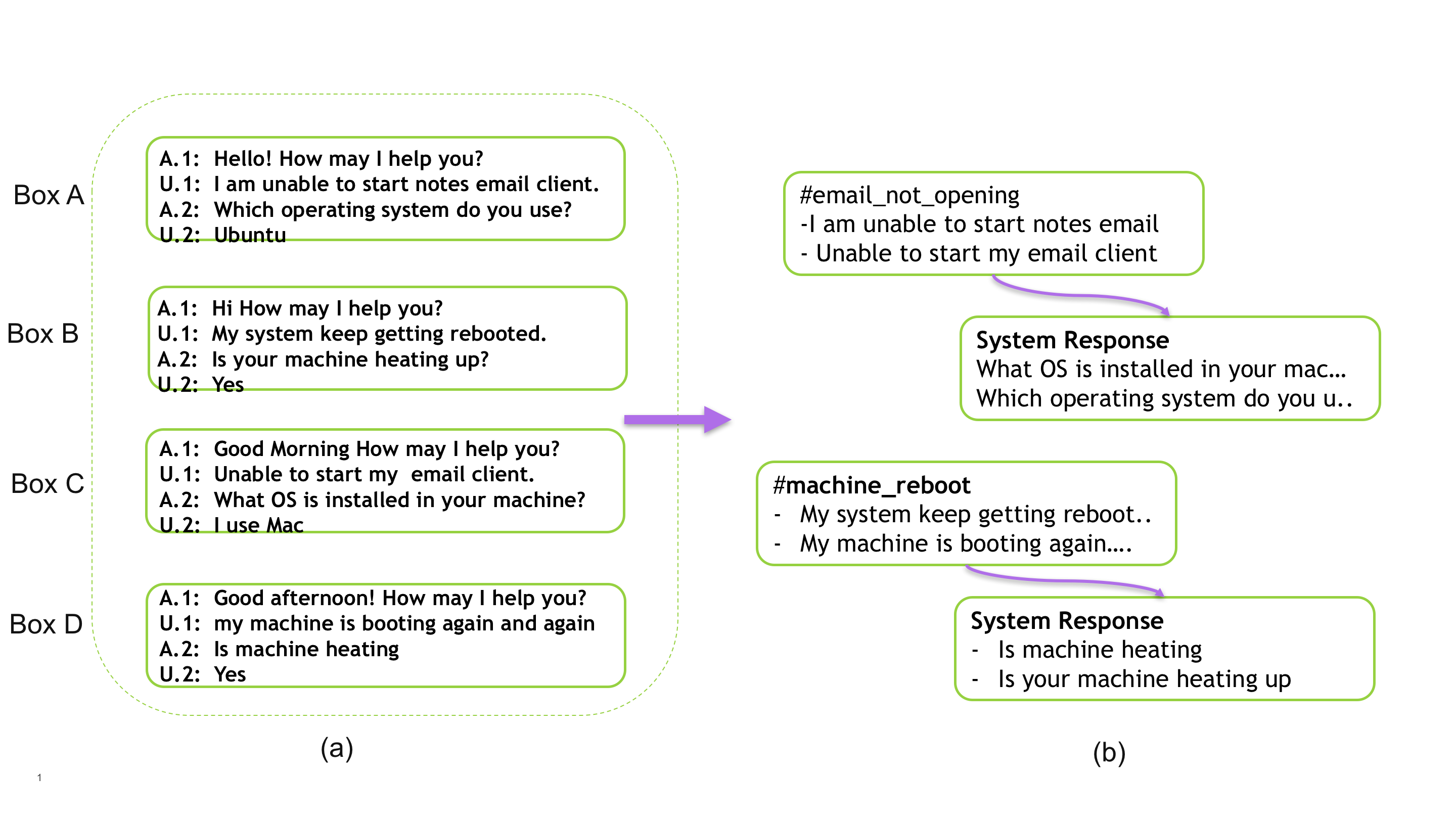}
  \captionsetup[figure]{skip=0pt}
  \caption{Some sample conversations and the obtained clusters}
  \label{fig:sample_conv}
  \end{center}
\end{figure*}

One approach to reduce the task of a dialog designer is to provide her with frequent user intents and possible corresponding system responses in a given domain. This can be done by analysing prior human to human conversations in the domain. Figure \ref{fig:sample_conv}(a) provides some example conversations in the technical support domain between users and agents.\\
In order to identify frequent user intents, one can use existing clustering algorithms to group together all the utterances from the users. Here each cluster would correspond to a new intent and each utterance in the cluster would correspond to an example for the intent. Similarly the agents utterances can be clustered to identify system responses. 
However, we argue that rather than treating user utterances and agents responses in an isolated manner,  there is merit in jointly clustering them. 
There is adjacency information of these utterances that can be utilized to identify better user intents and system responses. As an example, consider agent utterances A.2 in box A and A.2 in box C in Figure \ref{fig:sample_conv}(a). The utterances ``Which operating system do you use?" and ``What OS is installed in your machine" have no syntactic similarity and therefore may not be grouped together. However the fact that these utterances are adjacent to the similar user utterances ``I am unable to start notes email client" and ``Unable to start my email client"  provides some evidence that the agent utterances might be similar.  Similarly the user utterances  ``My system keeps getting rebooted" and ``Machine is booting time and again" ( box B and D in Figure \ref{fig:sample_conv}(a))- that are syntactically not similar - could be grouped together since the adjacent agent utterances, ``Is your machine heating up?" and ``Is the machine heating?" are similar. 

Joint clustering of user utterances and agent utterances allow us to align the user utterance clusters with agent utterance clusters. 
Figure \ref{fig:sample_conv}(b) shows some examples of user utterance  clusters and agent utterance clusters along with their alignments. Note that the user utterance clusters can be used by a dialog designer to specify intents, the agent utterance clusters can be used to create system responses and their alignment can be used to create part of the dialog flow.  

We propose two ways to take adjacency information into account. Firstly we propose a method called {\it SimCluster} for jointly or simultaneously clustering user utterances and agent utterances. SimCluster extends the K-means clustering method by incorporating additional penalty terms in the objective function that try to align the clusters together (described in Section \ref{sec:model_algo}).
The algorithm creates initial user utterance clusters as well as agent utterance clusters and then use bi-partite matching to get the best alignment across these clusters. 
 Minimizing the objective function pushes the cluster centroids to move  towards the centroids of the aligned clusters. 
The process  implicitly ensures that the similarity of adjacent agent utterances affect the grouping of  user utterances and conversely similarity of adjacent user utterances affect the grouping of agent utterances. In our second approach we use the information about neighbouring utterances for creating the vector representation of an utterance. For this we train a sequence to sequence model \cite{SVL14} to create the vectors (described in Section \ref{sec:expreal}). 

Our experiments described in  section \ref{sec:expreal} show that we achieve upto 10\% absolute improvement in F1 scores over standard K-means using SimCluster. Also we observe that clustering of customer utterances gains significantly by using the adjacency information of agent utterances whereas the gain in clustering quality of agent utterances is moderate. This is because the agent utterances typically follow similar syntactic constructs whereas customer utterances are more varied. Considering the agent utterances into account while clustering users utterances  is thus helpful. The organization of the rest of the paper is as follows.
In Section \ref{sec:related} we describe the related work. In Section \ref{sec:model_algo} we describe our problem formulation for clustering and the associated algorithm. Finally in sections \ref{sec:expsynth} and \ref{sec:expreal} we discuss our experiments on synthetic and real datasets respectively. 
\section{Related Work}
\label{sec:related}
The notion of adjacency pairs was introduced by Sacks et al \shortcite{SSE74} to formalize the structure of a dialog. 
Adjacency pairs have been used to analyze the semantics of the dialog in computational linguistics community \cite{PM00}. Clustering has been used for different tasks related to conversation. \cite{RCD10} considers the task of discovering dialog acts by clustering the raw utterances.   We aim to obtain the frequent adjacency pairs through clustering.\\
There have been several works regarding extensions of clustering to different scenarios such as:-
\begin{enumerate}
\item Co-clustering : Co-clustering considers the setting where data and features are clustered simultaneously. Dhillon \shortcite {Dhi01} considers a spectral graph theoretic approach to co-cluster documents and words simultaneously.  Dhillon, Mallela and Modha \shortcite{DMM03} consider an information theoretic formulation of co-clustering. 
 \item Multi task learning: Multi task learning considers task of learning from multiple domains simultaneously \cite{Car98}. Gu and Zhou \shortcite{GZ09} consider the  problem of multi task clustering wherein they cluster multiple domains simultaneously and utilize the relation of the domains to enhance clustering performance. Their model consists a reduced subspace in which the projection of vectors from the two domains have similar distribution. They then try to learn this common subspace and the clusters simultaneously. Our scenario differs from multi task learning since here the distributions across the domains tend to be different. (The domains being the possible utterances of user and agent).
\item Transfer learning considers the task of transfering the knowledge across similar tasks \cite{Dai09}.  Bhattacharya et al \shortcite{BGJV12} consider the task of clustering in the target domain using the given clusters in a source domain. They formulate the problem of minimizing a weighted sum of the energy function in the clustering, along with the energy of aligning the clusters of the two domains. This setting differs from ours since again the utterances in the two domains can be very different. Moreover unlike the  task of transfer learning we do not have clusters in any of the domains. However we do have information regarding the adjacency of utterances between the two domains.
\end{enumerate}


\section{The Proposed Approach}
\label{sec:model_algo}
In this section we describe our approach SimCluster that performs clustering in the two domains simultaneously and ensures that the generated clusters can be aligned with each other. 
We will describe the model in section \ref{sec:model} and the algorithm in Section \ref{sec:algorithm}.
\subsection{ Model}
\label{sec:model}
We consider a problem setting where we are given a collection of  pairs of consecutive utterances, with vector representations $\{x^{(i)},y^{(i)}\}_{i=1}^m$ where $x^{(i)}$s are in speaker 1's domain and $y^{(i)}$s are in speaker 2's domain. We need to simultaneously cluster the utterances in their  respective domains to minimize the variations within each domain and also ensure that the clusters for both domains are close together.\\
We denote the clusters for speaker 1's domain by $\{C_j^x\}_{j=1}^k$ with their respective means $\{\mu_j^x\}_{j=1}^k$. We denote the clusters assignments for $x^{(i)}$ by $ca^x(i)\in \{1,2,...,k\}$.\\
We denote the clusters for second speaker by $\{C_j^y\}_{j=1}^k$ with their respective means $\{\mu_j^y\}_{j=1}^k$. We denote the clusters assignments for $y^{(i)}$ by $Ca^y(i)\in \{1,2,...k\}$.
The usual energy function has the terms for distance of points from their corresponding cluster centroids. To be able to ensure that the clusters in each domain are similar, we also consider  an alignment between the centroids of the two domains. Since the semantic representations in the two domains are not comparable we consider a notion of \textbf{induced} centroids.\\
We define the induced centroids $\widetilde{\{\mu_j^x\}}_{j=1}^k$ as the arithmetic means of the points $\{x^{(i)}\}$s such that $y^{(i)}$'s have the same cluster assigned to them. Similarly, we define $\widetilde{\{\mu_j^y\}}_{j=1}^k$ as the arithmetic means of $\{y^{(i)}\}$s such that $x^{(i)}$s have the same cluster assigned to them. More formally, we define these induced centroids as:-
\[
\widetilde{\{\mu^x_j\}}=\frac{\sum_{i:Ca^y(i)=j} x^{(i)}}{\vert \{ i:Ca^y(i)=j\} \vert}\]
and
\[
\widetilde{\{\mu^y_j\}}=\frac{\sum_{i:Ca^x(i)=j} y^{(i)}}{\vert \{ i:Ca^x(i)=j\} \vert}\]
The alignment between these clusters given by the function $ma:[k] \mapsto [k]$, which is a bijective mapping from the cluster indices in speaker 1's domain to those in speaker 2's domain. Though there can be several choices for this alignment function, we consider this alignment to be a matching which \textbf{maximizes the sum of number of common indices} in the aligned clusters. More formally we define 
\[ N(j_1,j_2)=\vert\{ i:x^{(i)} \in C^x_{j_1} \text{ and } y^{(i)} \in C^y_{j_2}\} \vert
\] 
Then the matching $ma$ is defined to be the bijective function which maximizes $\sum_{j=1}^k N(j,ma(j))$. We consider a term in the cost function corresponding to the sum of distances between the original centroids and the matched induced centroids. Our overall cost function is now given by:-

\begin{align*}
J&=\alpha (\sum_{i=1}^m \|x^{(i)}-\mu^x_{Ca^{x}(i)}\|^2\\
&+\sum_{i=1}^m \|y^{(i)}-\mu^y_{Ca^y(i)}\|^2)\\
&+(1-\alpha)(\sum_{j=1}^k \|\mu^x_j-\widetilde{\mu^x_{ma(j)}}\|^2.\vert C^x_j \vert\\
&+\sum_{j=1}^k \|\mu^y_j-\widetilde{\mu^y_{ma^{-1}(j)}}\|^2 \vert C^y_j \vert)
\end{align*}

\begin{figure}[h!]
  \includegraphics[width=\linewidth]{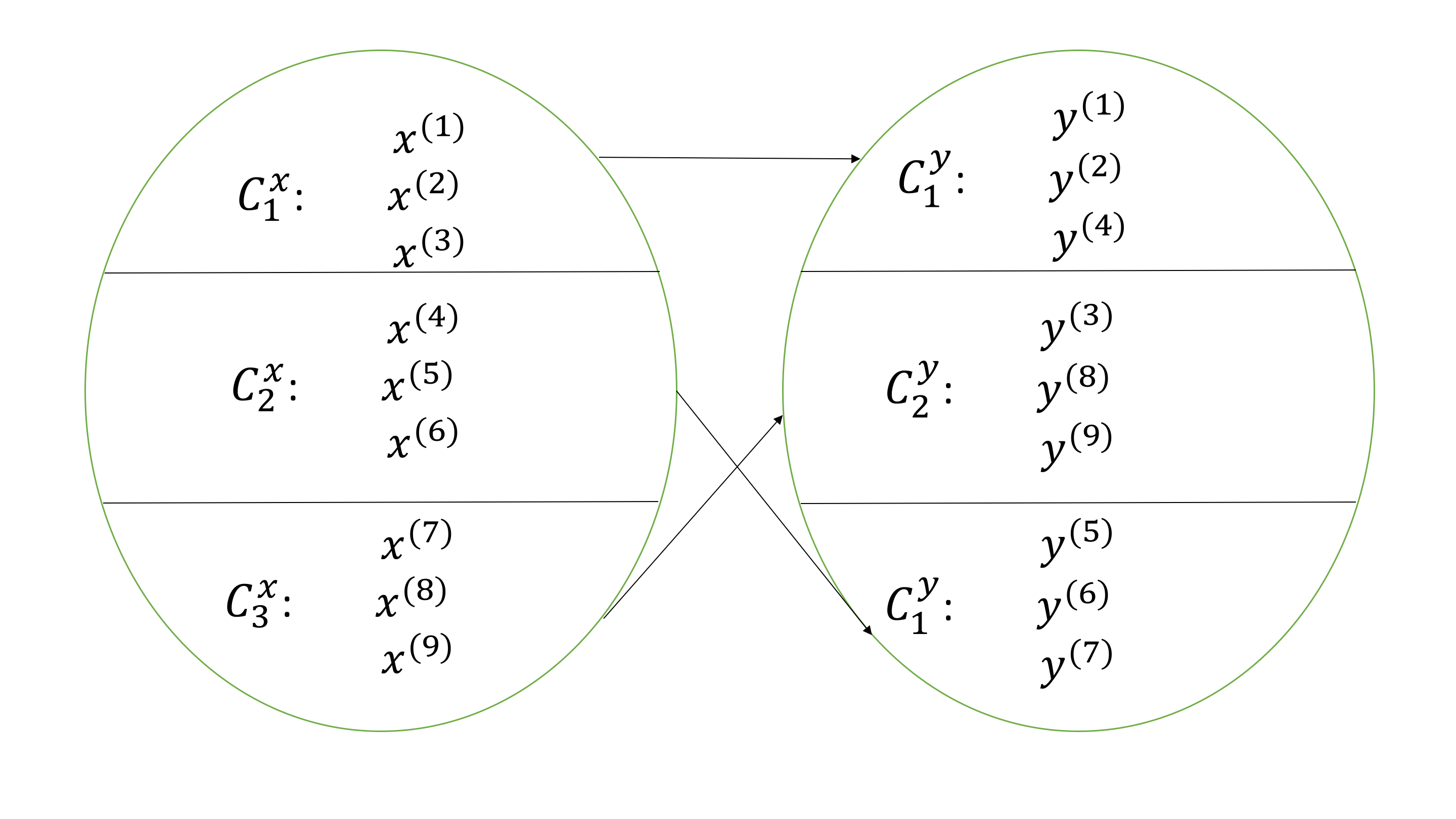}
  \caption{Sample clusters with matching}
  \label{fig:cluster_match}
\end{figure}

We explain the above definition via an example. Consider the clusters shown in Figure \ref{fig:cluster_match}. Here the $ma$ would match $C^x_1$ to $C^y_1$, $C^x_2$ to $C^y_3$ and $C^x_3$ to $C^y_2$, giving a match score of 6. Since $y^{(1)}$, $y^{(2)}$ and $y^{(4)}$ are present in the cluster $C^y_1$, $\widetilde{\mu^x_1}$ is given by $\frac{x^{(1)}+x^{(2)}+x^{(4)}}{3}$. Similarly \begin{align*}
\widetilde{\mu^x_2}&=\frac{x^{(3)}+x^{(8)}+x^{(9)}}{3}\\ \widetilde{\mu^x_3}&=\frac{x^{(5)}+x^{(6)}+x^{(7)}}{3}
\end{align*} In a similar manner, $\widetilde{\mu^y}$s can also be defined. Now the alignment terms are given by:-
\small
\begin{align*}
& \|\mu^x_1-\widetilde{\mu^x_1}\|^2 \vert C^x_1 \vert + \|\mu^x_2-\widetilde{\mu^x_3}\|^2 \vert C^x_2 \| + \|\mu^x_3-\widetilde{\mu^x_2}\|^2 \vert C^x_3 \vert +\\
&\|\mu^y_1-\widetilde{\mu^y_1}\|^2 \vert C^y_1 \vert + \|\mu^y_2-\widetilde{\mu^y_3}\|^2 \vert C^y_2 \| + \|\mu^y_3-\widetilde{\mu^y_2}\|^2 \vert C^y_3 \vert   
\end{align*}
\normalsize
%
\subsection{SimCluster Algorithm}
\label{sec:algorithm}
\begin{algorithm*}[]
\caption{SimCluster}\label{algo3}
\footnotesize
\begin{algorithmic}[1]
\Procedure{SimCluster}{Input: $\{(x^{(i)},y^{(i)})\}_{i=1}^m$,k (No. of cluster)}
\State{Output: A cluster assignment $Ca^x$ for   $x^{(i)}$s  and a cluster assignment $Ca^y$ for $y^{(i)}$s  }
\State{Initialize a set of centroids $\{\mu^x_j\}_{j=1}^k$  , and $\{\mu^y_j\}_{j=1}^k$} 
\State{Perform simple clustering for a few  iterations}
\Repeat 
\State{For each i, compute $Ca^x(i)$  as the index j among 1 to k which minimizes  $ \|x^{(i)}-\mu^x_j\|^2$.}
\State{Similarly , compute $Ca^y(i)$ as the index j' among 1 to k which minimizes $ \| y^{(i)}-\mu^y_{j'} \|^2 $.}
\State{Update the centroids, $\mu^x_j$ and $\mu^y_j$ as:-
\[ \mu^x_j=\alpha \left(\frac{\sum_{i: Ca^x(i)=j} x(i)}{\vert C^x_j \vert } \right)+(1-\alpha) \left(\frac{\sum_{i: Ca^y(i)=ma(j)} x(i)}{\vert C^y_j \vert } \right)\]
\hskip \algorithmicindent \hskip \algorithmicindent and \[ \mu^y_j=\alpha \left(\frac{\sum_{i: Ca^y(i)=j} y(i)}{\vert C^y_j \vert } \right)+(1-\alpha) \left(\frac{\sum_{i: Ca^x(i)=ma^{-1}(j)} y(i)}{\vert C^x_j \vert } \right)\]
}
\State{Perform a Hungarian matching between the cluster indices in the two domains with weights \\\hskip \algorithmicindent \hskip \algorithmicindent  N(j,j') on edges from index j to index j'.} 

\Until{convergence}
\EndProcedure
\end{algorithmic}
\end{algorithm*}
 \normalsize
To minimize the above energy term we adopt an approach similar to Lloyd's clustering algorithm \shortcite{Llo82} . 
We assume that we are given a set of initial seeds for the cluster centroids $\{\mu^x_j\}_{j=1}^k$ and $\{\mu^y_j\}_{j=1}^k$. 
We \textbf{repeat} the following steps iteratively:-
\begin{enumerate}
\item \label{algsumstep1}Minimize the energy with respect to cluster assignment keeping centroids unchanged. As in standard K-means algorithm, this is achieved by updating the cluster assignment, $Ca^x$ for each index i to be the cluster index j which minimizes $\| x^{(i)}-\mu^x_j\|^2$. Correspondingly for $Ca^y$, we pick the cluster index j' which minimizes $\| y^{(i)}-\mu^y_{j'}\|^2$.
\item \label{algsumstep2} Minimize the energy with respect to the centroids keeping cluster assignment unchanged. To achieve this step we need to minimize the energy function with respect to the centroids $\mu^x_j$ and $\mu^y_j$. This is achieved by setting $\nabla_{\mu^x_j}J=0$ for each j and $\nabla_{\mu^y_j}J=0$ for each j.\\
Setting $\nabla_{\mu^x_j}J=0$, we obtain 
\small
\[ \mu^x_j=\alpha \left(\frac{\sum_{i: Ca^x(i)=j} x(i)}{\vert C^x_j \vert } \right)+(1-\alpha) \widetilde{\mu^x_{ma(j)}}\]
\normalsize
or equivalently

\begin{align*}
 \mu^x_j&=\alpha \left( \frac{\sum_{i: Ca^x(i)=j} x(i)}{\vert C^x_j \vert } \right) \\&+(1-\alpha)\left (\frac{\sum_{i: Ca^y(i)=ma(j)} x(i)}{\vert C^y_j \vert } \right )
 \end{align*}
Similarly, setting $\nabla_{\mu^y_j}J=0$, we obtain 
\begin{align*}
 \mu^y_j&=\alpha \left(\frac{\sum_{i: Ca^y(i)=j} y(i)}{\vert C^y_j \vert } \right)\\&+(1-\alpha) \left(\frac{\sum_{i: Ca^x(i)=ma^{-1}(j)} y(i)}{\vert C^x_j \vert } \right)
 \end{align*}
\item Finally we update the matching between the clusters. To do so, we need to find a bipartite matching match on the cluster indices so as to maximize $\sum_{j=1}^k N(j,ma(j))$. We use Hungarian algorithm \cite{Kuh95} to perform the same i.e. we define a bipartite graph with vertices consisting of cluster indices in the two domains. There is an edge from vertex representing cluster indices j (in domain 1) and j' in domain 2, with weight N(j,j'). We  find a maximum weight bipartite matching in this graph. 
\end{enumerate}
Similar to Lloyd's algorithm, each step of the above algorithm decreases the cost function. This ensures that the algorithm achieves a local minima of the cost function if it converges.
See Algorithm \ref{algo3} for a formal description of the approach. The centroid update step of the above algorithm also has an intuitive explanation i.e. we are slightly moving away the centroid towards the matched induced centroid. This is consistent with our goal of aligning the clusters together in the two domains.  
\begin{table*}[]
\centering
\begin{tabular}{|l|l|l|l|l|}\hline
 & \multicolumn{2}{l|}{Domain 1} & \multicolumn{2}{l|}{Domain 2} \\ \hline
 & F1-score & ARI & F1-score & ARI \\ \hline
K-means  & 0.412 & 0.176 & 0.417 & 0.180 \\ \hline
SimCluster & 0.442 & 0.203 & 0.441 & 0.204 \\ \hline
\end{tabular}
\caption{Performance of SimCluster versus K-means clustering on synthetic dataset}
\label{table:synth_results}
\end{table*}
\subsection{Alignment}
\label{sec:aligning}
The algorithm above maintains a mapping between the clusters in each speaker's domain. This mapping serves to give us the alignment between the clusters required to provide a corresponding response for a given user intent.
\section{Experiments on Synthetic Dataset}
\label{sec:expsynth}
We performed experiments on synthetically generated dataset since it gives us a better control over the distribution of the data. 
Specifically we compared the gains obtained using our approach versus the variance of the distribution.
We created dataset from the following generative process.
\begin{algorithm}[H]
\caption{Generative Process}\label{algo3}
\begin{algorithmic}[1]
\Procedure{Generate data}{}

\State{ Pick k points $\{\mu_x^{(i)}\}_{i=1}^k$ as domain -1 means and a corresponding set of k points   $\{\mu_y^{(i)}\}_{i=1}^k$ as domain-2 means, and covariance matrices $\Sigma_x and \Sigma_y$}

\For{ iter$\gets 1$ upto num$\_$samples}
\State{Sample class $ \sim U\{1,2...k\}$}
\State{Sample $q\sim \mathcal{N}(\mu_x^{class},\Sigma_x)$}
\State{Sample $a\sim \mathcal{N}(\mu_y^{class},\Sigma_y)$}
\State{Add q and a so sampled to the list of q,a pairs}
\EndFor
\EndProcedure
\end{algorithmic}
\end{algorithm}
We generated the dataset from the above sampling process with means selected on a 2 dimensional grid of size $3\times 3$ with variance set as $\frac{1}{2}$ in each dimension.
10000 sample points were generated. 
The parameter $\alpha$ of the above algorithm was set to 0.5
and k was set to 9 (since the points could be generated from one of the 9 gaussians with   centroids on a $3\times 3$ grid).\\
We compared the results with simple K-means clustering with k set to 9.
For each of these, the initialization of means was done using $D^2$ sampling approach \cite{AV07}.
\subsection{Evaluation and Results}
To evaluate the clusters we computed the following metrics 
\begin{enumerate}

\item ARI (Adjusted Rand Index): Standard Rand Index is a metric used to check the clustering quality against a given standard set of clusters by comparing the pairwise clustering decisions. It is defined as $\frac{a+b}{a+b+c+d}$, where a is the number of true positive pairs, b is the number of true negative pairs, c is the number of false positive pairs and d is the number of false negative pairs. Adjusted rand index corrects the standard rand index for chance and is defined as $\frac{\text{ Index } - \text{ Expected index  }}{\text{ Max Index }-\text{ Expected index } }$ \cite{Ran71}. \\
We compute ARI score for both the source clusters as well as the target clusters.
\item F1 scores:  We also report F1 scores for the pairwise clustering decisions. In the above notation we considered the pair-precision as $\frac{a}{a+c}$ and recall as $\frac{a}{a+d}$. The F1 measure is the Harmonic mean given as $\frac{2PR}{P+R}$.
\end{enumerate}

We used the gaussian index from which an utterance pair was generated as the ground truth label, which served to provide ground truth clusters for computation of the above evaluation metrics.
Table \ref{table:synth_results} shows a comparison of the results on SimCluster versus K-means algorithm. Here our SimCluster algorithm improves the F1-scores from 0.412 and 0.417 in the two domains to 0.442 and 0.441. The ARI scores also improve from 0.176 and 0.180 to 0.203 and 0.204.
%
%
\begin{table*}[]
\centering
\begin{tabular}{|l|l|l|l|l|}\hline
 & \multicolumn{2}{l|}{Customer} & \multicolumn{2}{l|}{Agent} \\ \hline
 & F1-score & ARI & F1-score & ARI \\ \hline
K-means (Doc2Vec) & 0.787 & 0.150 & 0.783 & 0.136 \\ \hline
SimCluster (Doc2Vec) & 0.88 & 0.19 & 0.887 & 0.192 \\ \hline
K-means  (Seq2Seq)  & 0.830 & 0.159 & 0.900 & 0.218 \\ \hline
SimCluster (Seq2Seq) & 0.860 & 0.181 & 0.916 & 0.218 \\\hline
\end{tabular}
\caption{Performance of SimCluster versus  K-means clustering on both Doc2Vec as well as seq2seq based vectors}
\label{table:combined_results}
\end{table*}
\subsubsection{Variation with variance}
\begin{figure}[h!]
  \includegraphics[width=\linewidth]{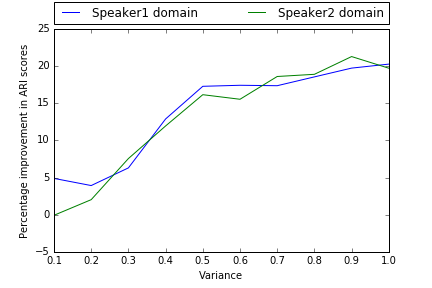}
  \caption{Improvement in ARI figures achieved by SimCluster versus variance }
  \label{fig:ari_var}
\end{figure}

\begin{figure}[h!]
  \includegraphics[width=\linewidth]{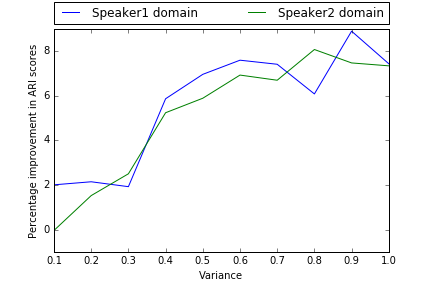}
  \caption{Variation of Improvement in F1 score figures achieved by SimCluster versus variance}
  \label{fig:f1_var}
\end{figure}

We also performed experiments to see how  the performance of SimCluster is affected by the variance in the cluster (controlled by the generative process in Algorithm \ref{algo3}). Intuitively we expect SimCluster to obtain an advantage over simple K-means when variance is larger. This is because at larger variance, the data points  are more likely to be generated away from the centroid due to which they might be clustered incorrectly with the points from neighbouring cluster. However if the corresponding point from the other domain is generated closer to the centroid, it might help in clustering the given data point correctly. We performed these experiments with points generated from Algorithm \ref{algo3} at differet values of variance. We  generated the points with centroids located on a grid of size $3\times 3$ in each domain. The value of k was set to 9. The experiment was repeated for each value of variance between 0.1 to 1.0 in the intervals of 0.1.
Figures \ref{fig:ari_var} and \ref{fig:f1_var} show the percentage improvement on ARI score and F1 score respectively achieved by SimCluster (over K-means) versus variance.

\section{Experiments on Real Dataset} 
\label{sec:expreal}
\subsection{Description and preprocessing  of dataset}
\begin{table*}[h!]
\tiny
\centering
\begin{tabular}{|l|l|}
\hline
Clusters in user domain & Clusters in agent domain\\\hline
\begin{tabular}{@{}l@{}}no refund got for the refund request made on 12 ...\\
11 days \& counting on a refund . was promised 3-5 days ... \\
yes i contacted my bank 2 days ... there is no sign of an amazon refund processing .\\
\textbf{ @amazon refer screen shot ..When will i get my refund ?...}\\
\textbf{ ... I have to wait 3-5 business days for Amazon to refund me money ...}
\end{tabular} &\begin{tabular}{@{}l@{}}... You can view your order refund status here ...\\
It can take up to 10 business days ... the refund has been processed\\
\textbf{...  if you have received the refund reference number then ... contact  support team  ...}\\
\textbf{Refunds typically take 5-7 days to show on your account ...}\\
\textbf{... As soon as the product reaches the shipper the refund will be initiated ... }
\end{tabular}
\\\hline
\begin{tabular}{@{}l@{}}...my package is late so why did I get prime ?\\ 
Paid \$ 20 + in shipping for next day delivery yesterday but ...  even tho I have prime\\
\textbf{ my order wasnt delivered yesterday ... why am I pay for prime ?}\\
\textbf{do you bother to let anyone know ... What is the point of prime ?}\\
\textbf{I'm a Amazon prime member . You promised me 2day delivery ...}
\end{tabular}& 
\begin{tabular}{@{}l@{}} 
I'm sorry to see it's late ... \\
I'm sorry it arrived late, but glad you did receive it .\\
\textbf{I'm glad to hear it was delivered , but I'm sorry it was a day late ...}\\
\textbf{I'm sorry your order is late ! When you contacted us...}\\
\end{tabular}\\\hline
%
%
\end{tabular}   
\caption{Sample clusters in user and agent domains. Utterances in bold are those which were not in the given cluster using K-means, but could be correctly classified with the cluster using SimCluster}
\label{table:samp_utter}
\end{table*}

We have experimented on a dataset containing Twitter conversations between customers and Amazon help. The dataset consisted of 92130 conversations between customers and amazon help. We considered the conversations with exactly two speakers Amazon Help and a customer. Consecutive utterances by the same speaker were concatenated and considered as a single utterance. From these we extracted adjacency pairs of the form of a customer utterance followed by an agent (Amazon Help) utterance. We then selected the utterance pairs from  8 different categories, like late delivery of item, refund, unable to sign into the account, replacement of item, claim of warranty, tracking delivery information etc. A total of 1944 utterance pairs were selected.
\\
To create the vector representation we had used two distinct approaches:-
\begin{enumerate}
\item  Paragraph to vector approach (Doc2Vec) by Le and Mikolov \shortcite{LM14}. Here we trained the vectors using distributed memory algorithm and trained for 40 iterations. A window size of 4 was used.
\item We also trained the vectors using sequence to sequence approach \cite{SVL14}, on the Twitter dataset where we considered the task of predicting the reply of Amazon Help for customer's query and vice versa.\\
The encoded vector from the input sequence forms the corresponding vector representation. For the task of generating the agent's response for customer utterance the encoding from the input sequence (in the trained model) forms the vector representation for the customer utterance. Similarly for the task of generating the previous customer utterance from the agent's response, the intermediate encoding forms the vector representation for the agent utterance. We used an LSTM based 3-layered sequence to sequence model with attention for this task. 
\end{enumerate}
We ran the K-means clustering algorithm for 5 iterations followed by our SimCluster algorithm for 30 iterations to form clusters in both the (customer and agent) domains. The hyper parameter($\alpha$) is chosen based on a validation set. We varied  the value of $\alpha$ from 0.5 to 1.0 at intervals of 0.025. The initialization of centroids was performed using $D^2$ sampling approach \cite{AV07}. 

\subsection{Results}
For the clusters so obtained we have computed F1 and ARI measures as before and compared with the K-means approach. We used the partitioning formed by the 8 categories (from which the utterance pairs were selected) as the ground truth clustering.\\
Table \ref{table:combined_results} summarizes the results. We observe that for K-means algorithm, the vectors generated from sequence to sequence model perform better than the vectors generated using paragraph to vector for both the domains. This is expected as the vectors generated from sequence to sequence model encode some adjacency information as well. We further observe that the SimCluster approach performs better than the K-means approach for both the vector representations. It improves the F1-scores for Doc2Vec representation from 0.787 and 0.783 to 0.88 and 0.887 in the two domains. Also the F1-scores on Seq2Seq based representation improve from 0.83 and 0.9 to 0.86 and 0.916 using SimCluster. However the gains are much more in case of Doc2Vec representations than Seq2Seq representations since Doc2Vec did not have any information from the other domain where as some amount of this information is already captured by Seq2Seq representation. 
Moreover it is the clustering of customer utterances which is likely to see an improvement. This is because agent utterances tends to follow a generic pattern while customer utterances tend to be more varied. Considering agent utterances while generating clusters in the user domain thus tends to be more helpful than the other way round. 

Table \ref{table:samp_utter} shows qualitative results on the same dataset. Column 1 and 2 consists of clusters of utterances in customer domain and agent domain respectively. The utterances with usual font are  representative utterances from clusters obtained through K-means clustering. The utterances in bold face indicate the similar utterances which were incorrectly classified in different clusters using K-means but were correctly classified together with the utterances  by SimCluster  algorithm.

\section{Conclusions}
One of the first steps to automate the construction of conversational systems could be to identify the frequent user utterances and their corresponding system responses. In this paper we proposed an approach to compute these groups of utterances by clustering the utterances in both the domains  using our novel SimCluster algorithm which seeks to simultaneously cluster the utterances and align the utterances in two domains. Through our experiments on synthetically generated datset we have shown that SimCluster has more advantage over K-means on datasets with larger variance. Our technique improves upon the ARI and F1 scores on a real dataset containing Twitter conversations. 
\section*{Acknowledgments}
We thank Dr. David Nahamoo (CTO, Speech Technology and Fellow IBM Research ) for his valuable guidance and feedback. We also acknowledge the anonymous reviewers of IJCNLP 2017 for  their comments. 
\bibliography{ijcnlp2017}
\bibliographystyle{ijcnlp2017}
\end{document}